\documentclass[conference]{IEEEtran}
\IEEEoverridecommandlockouts
\usepackage{cite}
\usepackage{amsmath,amssymb,amsfonts}
\usepackage{algorithmic}
\usepackage{graphicx}
\usepackage{textcomp}
\usepackage{xcolor}
\usepackage{multirow}
\def\BibTeX{{\rm B\kern-.05em{\sc i\kern-.025em b}\kern-.08em
    T\kern-.1667em\lower.7ex\hbox{E}\kern-.125emX}}
\begin{document}

\title{FANeRV: Frequency Separation and Augmentation based Neural Representation for Video}

\author{Li~Yu, ~Zhihui~Li, ~Chao~Yao, ~Jimin~Xiao, ~and~Moncef~Gabbouj,~\IEEEmembership{Fellow,~IEEE}}

\maketitle

\begin{abstract}Neural representations for video (NeRV) have gained considerable attention for their strong performance across various video tasks. However, existing NeRV methods often struggle to capture fine spatial details, resulting in vague reconstructions. In this paper, we present a Frequency Separation and Augmentation based Neural Representation for video (FANeRV), which addresses these limitations with its core Wavelet Frequency Upgrade Block. This block explicitly separates input frames into high and low-frequency components using discrete wavelet transform, followed by targeted enhancement using specialized modules. A specially designed gated network effectively fuses these frequency components for optimal reconstruction. Additionally, convolutional residual enhancement blocks are integrated into the later stages of the network to balance parameter distribution and improve the restoration of high-frequency details. Experimental results demonstrate that FANeRV significantly improves reconstruction performance and excels in multiple tasks, including video compression, inpainting, and interpolation, outperforming existing NeRV methods.
\end{abstract}


\section{Introduction}
\label{sec:intro}

Implicit Neural Representations (INRs) have attracted significant attention for their remarkable ability to accurately represent diverse signals\cite{nerf,siren,3d1} . 
NeRV \cite{nerv} demonstrates the potential of INR in video representation. NeRV introduces a frame-based implicit representation that utilizes convolutional layers to map frame indices directly to video frames. It addresses issues such as slow training and poor reconstruction quality in pixel-based implicit representations \cite{coin}, while demonstrating its potential across a variety of tasks, including video compression, interpolation, and inpainting. Building on this paradigm, subsequent works \cite{E-nerv,hnerv,boost} have focused on designing more efficient network structures and embedding methods to further improve video reconstruction quality. 

Despite these advancements, neural networks are constrained by spectral bias \cite{spectralbias}, which limits their ability to effectively capture fine spatial details, resulting in blurry video frame reconstruction. To address this issue, some studies \cite{vqnerv,qsnerv} encode high-frequency embeddings to improve detail recovery. However, these additional feature embeddings increase both network parameters and computational overhead. Other approaches \cite{snerv,nerv++} adopt residual concatenation strategies, integrating low-resolution features from earlier stages to accelerate high-frequency learning and enhance reconstruction quality. However, improper upsampling alignment in these methods can cause frequency aliasing, impeding further performance improvements. Additionally, existing methods do not explicitly separate and learn high and low-frequency components. Instead, they employ a uniform approach to all frequency components, making it challenging to emphasize high-frequency details while retaining the richness of low-frequency features.




To address these challenges, we propose the Frequency Separation and Augmentation based Neural Representation for video (FANeRV), designed to enhance the recovery of high-frequency details in frame reconstruction. Our method employs discrete wavelet transforms to explicitly decompose input features into high and low-frequency components, followed by applying targeted enhancement modules for each. Additionally, low-resolution features are fused with low-frequency subband to reinforce gradient flow and direct the network’s focus toward high-frequency details, enabling the precise capture of subtle spatial variations. To this end, we propose a Wavelet Frequency Upgrade Block (WFUB), which comprises two key components: the Frequency Separation Feature Boosting (FSFB) module and the Time-Modulated Gated Feed-Forward Network (TGFN). The FSFB module contains a multi-resolution deep feature modulation branch for capturing non-local dependencies, and a small-kernel residual branch to model local details features. The synergy of these sub-branches enables the FSFB module to effectively extract and integrate features across high and low-frequency components.
Following this, the TGFN component refines the fused features across spatial and channel dimensions, focusing the network’s attention on regions with rich information. Furthermore, we integrate convolutional residual enhancement blocks in the later stages of the network to optimize parameter distribution and enhance precision in reconstructing high-frequency details. 
\begin{figure*}[htbp]
\centering
\includegraphics[width=0.85\linewidth]{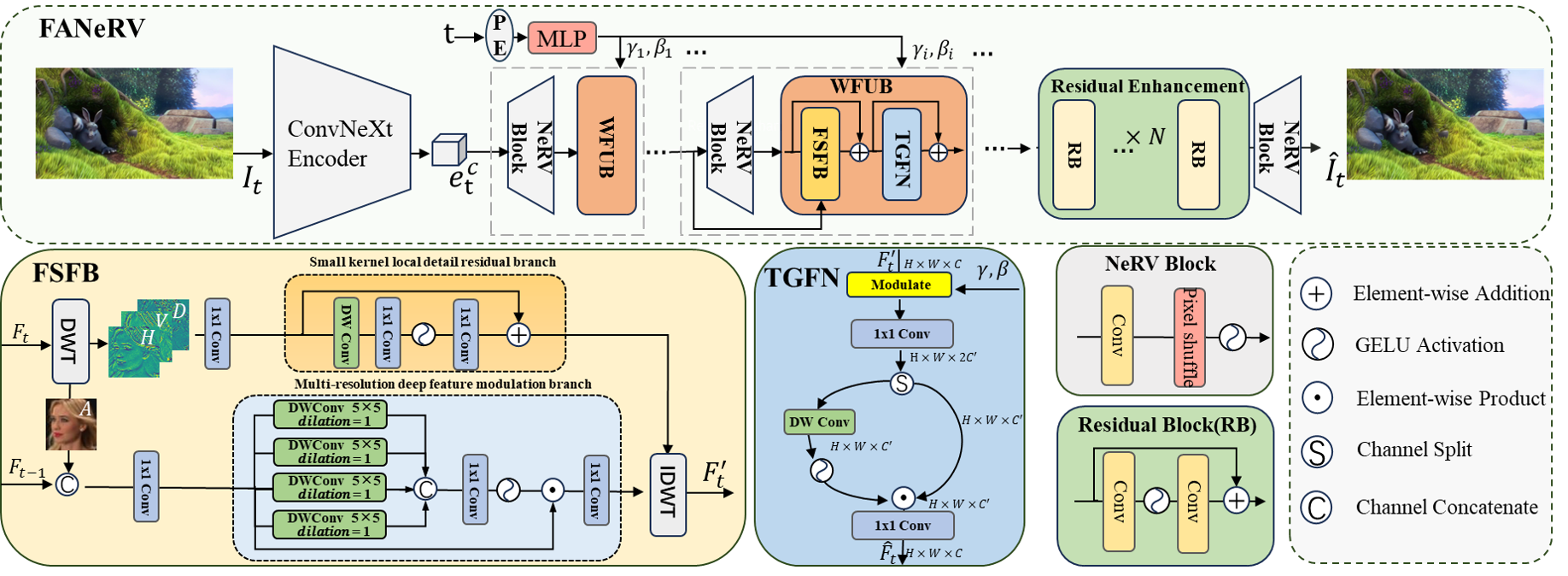}
\caption{Architecture of the proposed FANeRV. FANeRV integrates the Wavelet Frequency Upgrade Block (WFUB) and the convolutional residual enhancement block to enhance detail reconstruction. The WFUB comprises a Separation Feature Boosting (FSFB) module and a Time-Modulated Gated Feed-Forward Network (TGFN).
}
\label{model}
\end{figure*} 

Our primary contributions are as follows:
\begin{itemize}
\item We introduce FANeRV, which employs wavelet transforms to decompose high and low-frequency components, and designs enhancement fusion strategies tailored to their characteristics. By integrating lower-resolution features, FANeRV efficiently leverages multi-scale and multi-frequency information to improve reconstruction quality and detail accuracy.

\item  We develop the FSFB module, which consists of a low-frequency branch for capturing non-local dependencies and a high-frequency branch for modeling local textures. To improve feature integration, we incorporate the TGFN to effectively refine the fused features. Additionally, convolutional residual enhancement blocks are integrated into the network's later stages to optimize parameter distribution and improve its capacity for reconstructing high-frequency details.

\item We conduct comprehensive experiments, demonstrating that FANeRV achieves significant performance improvements across multiple tasks, including video regression, video compression, image inpainting, and interpolation, outperforming a wide range of baseline models.
\end{itemize}

\section{Proposed method}

We begin by presenting an overview of our method pipeline in Section \ref{sec:II-A}. Next, we delve into the details of the proposed representation network in Sections \ref{sec:II-B} and \ref{sec:II-C}. Finally, we outline the loss function in Section \ref{sec:II-D}.

\subsection{Overview.}
\label{sec:II-A}
Fig. \ref{model} illustrates the architecture of the proposed Frequency Separation and Augmentation based Neural Representation for video (FANeRV). FANeRV employs a hybrid approach\cite{hnerv} for video representation, comprising content-related feature embeddings and decoder parameters. Specifically, given a video frame sequence \( V = \{ I_0, I_1, \dots, I_{T-1} \} \), where \( I_t \in \mathbb{R}^{H \times W \times C} \) represents the frame at timestamp \( t \) with height \( H \), width \( W \), and channels \( C \). The input frames \( I_t \) are initially downsampled to generate compact, content-related feature embeddings \( e_t^c \) via a ConvNext encoder \( E \), which consists of stacked ConvNext blocks \cite{convnet}. These embeddings \( e_t^c \) are then processed by the decoder, which is composed of a series of NeRV blocks \cite{hnerv} and Wavelet Frequency Upgrade Blocks (WFUB), to perform feature upsampling and refinement. In parallel, a temporal modulation network injects time information into the intermediate features at each stage, ensuring effective alignment during decoding. Finally, the refined features are passed through stacked convolutional residual enhancement blocks, yielding the final high-quality reconstruction.
\subsection{Wavelet Frequency Upgrade Block}
\label{sec:II-B}
In image restoration and reconstruction tasks, preserving and enhancing high-frequency features is critical for achieving high visual fidelity. Research has demonstrated that independently amplifying high and low-frequency features is an effective strategy \cite{crnet,asymmetric,afftransformer}. Inspired by this, we proposed the WFUB to explicitly separate and strengthen high and low-frequency components. It consists of the Frequency Separation Feature Boosting (FSFB) module and Time-Modulated Gated Feed-Forward Network (TGFN). Specifically, given an input feature \( F_t \in \mathbb{R}^{H \times W \times C} \), where \( H \times W \) denotes the spatial dimensions and \( C \) represents the number of channels, we employ the Haar wavelet transform \cite{harr} to decompose each \( F_t \) into high and low-frequency components, resulting in four subbands:
\[
\{A_{LL}^t, H_{LR}^t, V_{RL}^t, D_{RR}^t\} = \text{DWT}(F_t)
\]
Here, \( A_{LL}^t \) represents the low-frequency subband containing global structural information, while \( H_{LR}^t \), \( V_{RL}^t \), and \( D_{RR}^t \) are high-frequency subbands capturing horizontal, vertical, and diagonal texture details. Each subband has dimensions \( \mathbb{R}^{H/2 \times W/2 \times C} \). Additionally, to utilize low-resolution features \( F_{t}^{i-1} \) from the previous stage, which primarily contain low-frequency information, we concatenate \( A_{LL}^t \) with \( F_{t}^{i-1} \) to form enhanced low-frequency components. 

After obtaining the explicitly extracted high-frequency
components and enhanced low-frequency components, we apply a \(1 \times 1\) convolution to transform them into initial high and low-frequency features. Next, we apply the proposed multi-resolution deep feature modulation branch $\mathcal{LB}$ to capture global dependencies across scales and the small-kernel residual branch $\mathcal{HB}$ to enhance local texture details. These branches work together to model both local and global features, enabling precise reconstruction. This process is defined as: 
\begin{align}
\{\hat{F_l^t}, \hat{F_h^t}\} = & \ \mathcal{LB}\left( \text{Conv1x1}(\text{Concat}(A_{LL}^t, F_{t}^{i-1})) \right), \notag \\
& \ \mathcal{HB}\left( \text{Conv1x1}((H_{LR}^t, V_{RL}^t, D_{RR}^t)) \right)
\end{align}
where \( \hat{F_l^t} \in \mathbb{R}^{H/2 \times W/2 \times C} \) and \( \hat{F_h^t} \in \mathbb{R}^{H/2 \times W/2 \times 3C} \) are the refined low-frequency and high-frequency features, respectively.
Next, we employ the inverse discrete wavelet transform (IDWT) to merge \( \hat{F_l^t} \) and \(\hat{F_h^t} \), producing an initial fused feature \( F_t' \). To further integrate these features and incorporate temporal embedding at each stage, we utilize TGFN to dynamically select the most representative features, refining the fused and time-modulated output. The process is defined as:

\begin{equation}
F_t' = \text{IDWT}(F_l^t, F_h^t) + F_t,
\end{equation}
\begin{equation}
\gamma_i, \beta_i = \text{MLP}(\text{PE}(t)),
\end{equation}
\begin{equation}
\hat{F_t} = \text{TGFN}(F_t'|\gamma_i, \beta_i) + F_t'.
\end{equation}

 where \( \text{PE} \) is the positional encoding function \cite{nerf}, \( \text{MLP} \) is a small multi-layer perceptron, and \( \gamma_i \) and \( \beta_i \) are affine parameters derived from temporal embeddings \( t \). Finally, \( \hat{F_t} \) represents the final output of the WFUB.

\subsubsection{Multi-Resolution Deep Feature Modulation Branch}
 Low-frequency features are essential for capturing structural and global characteristics in images. Exploring non-local feature interactions is crucial for their effective restoration. While existing methods use self-attention mechanisms or large kernel convolutions to explore non-local information and achieve strong reconstruction performance, they are often computationally expensive and are parameter-intensive. To address these challenges, we propose a lightweight approach for learning long-range dependencies from multi-scale feature representations. As shown in Fig. \ref{model}, we employ parallel depthwise separable dilated convolutions with varying kernel sizes and dilation rates to construct multi-scale feature maps \(f_l^{j}\) that capture low-frequency features at different scales. Specifically, given an input feature \( F_l^t \in \mathbb{R}^{H \times W \times C} \), where \( H \) and \( W \) are the spatial dimensions and \( C \) is the number of channels, this procedure can be expressed as:

\begin{equation}
    f_l^{j} = \text{DWConv}_{k_j \times k_j}(F_l^t, d_j), \quad 0 \leq j \leq 3
\end{equation}

where \(\text{DWConv}_{k_j \times k_j}(\cdot)\) denotes a depthwise separable convolution with kernel size \( k_j \) and dilation rate \( d_j \), specifically set as \( k_j = [5, 7, 9, 11] \) and \( d_j = [1, 2, 3, 4] \).
To aggregate features from multi-scale low-frequency representations, we incorporate a gating mechanism to adaptively select the most representative features. we concatenat these features and then apply a \(1 \times 1\) convolutional layer to produce a refined global feature representation that captures multi-order contexts. The GELU activation function \cite{hendrycks2016gaussian} is then applied for nonlinear normalization, generating an attention map. This attention map is element-wise multiplied with the input features, enabling adaptive feature modulation that emphasizes critical features while dynamically adjusting the contribution of each scale. Finally, another \(1 \times 1\) convolutional layer models the inter-channel relationships, generating the final output \( \hat{F_l^i} \):

\begin{equation}
    S = \sigma \left( \text{Conv}_{1 \times 1} \left( \text{Concat}([f_l^{0}, f_l^{1}, f_l^{2}, f_l^{3}]) \right) \right)
\end{equation}
\begin{equation}
    \hat{F_l^t} = \text{Conv}_{1 \times 1} (S \odot F_l^t)
\end{equation}
where \( \sigma(\cdot) \) represents the GELU activation function, and \( \odot \) denotes element-wise product.
This approach maintains computational efficiency while improving the network's capability to handle global structural information in images.

\subsubsection{Small Kernel Local Detail Residual Branch}
High-frequency features represent essential local details in images, crucial for high-quality reconstruction.
To restore finer details, it is necessary to use smaller receptive fields that focus more precisely on local image information. Thus, we propose the Small-Kernel Local Detail Residual Block. Small convolution kernels enhance the focus on local details, while residual connections effectively capture high-frequency information. Specifically, the \(3 \times 3\) depthwise convolution encodes local information \( f_h \) from the input \( F_h^t \) while expanding the channel count for enhanced channel mixing. Then, we use two \(1 \times 1\) convolutions to further explore channel information and generate the enhanced local feature \( \hat{F_h^t} \), which is achieved by:
\begin{equation}
    f_h = \text{Conv}_{1 \times 1} \left( \text{DWConv}_{3 \times 3} (F_h^t) \right)
\end{equation}
\begin{equation}
    \hat{F_h^t} = \text{Conv}_{1 \times 1} \left( \sigma(f_h) \right)+F_h^t
\end{equation}
where \( \sigma(\cdot) \) denotes the GELU activation function. 

\subsubsection{Time-Modulated Gated Feed-Forward Network}
To effectively fuse low-frequency and high-frequency features, we introduce a gating network that dynamically selects and integrates features, thereby reducing redundant processing. Moreover, the integration of temporal context at each stage enhances regression performance, accelerates model convergence, and improves the overall quality of the results \cite{E-nerv}. To achieve this, temporal information is embedded into the intermediate features prior to the application of convolutional gating. Specifically, the input features \(F_i'\) are first linearly transformed using modulation parameters \(\gamma_i\) and \(\beta_i\) derived from temporal embeddings:

\begin{equation}
f_t^{\prime \text{modulate}} = F_t' \odot (1+\gamma_i) + \beta_i
\end{equation}

A \(1 \times 1\) convolution with GELU activation is then applied for cross-channel interaction in the expanded hidden space. The output is split into two components, \( a \) and \( \mathbf{x} \). Component \( a \) undergoes a \(3 \times 3\) depthwise convolution to capture local spatial patterns, followed by a GELU non-linearity to estimate the attention map. This estimated attention is then used to adaptively modulate \( \mathbf{x} \) through an element-wise product.
 Finally, a \(1 \times 1\) convolution mixes features and reduces channel dimensions to match the input. The process is described by:
\begin{equation}
f_a, f_x = \text{Split}(\text{Conv}_{1 \times 1}(f_t^{\prime \text{modulate}}))
\end{equation}
\begin{equation}
{\hat{F_t}} = \text{Conv}_{1 \times 1}(f_x \odot \sigma(\text{DWConv}_{3 \times 3}(f_a)))
\end{equation}

where \( \sigma(\cdot) \) represents GELU activation, and \( \odot \) denotes element-wise product.

\subsection{Convolutional Residual Enhancement Block}
\label{sec:II-C}
Since image resolution is typically increased progressively during up-sampling, previous works\cite{nerv,hnerv,boost,E-nerv} often reduce the number of channels after each stage to maintain a compact network size and acceptable computational cost. However, this approach can lead to a reduction in the number of parameters in later stages, which may hinder the model's ability to recover fine image details. To address this issue, we integrate convolutional residual enhancement block (CREB) after the final WFUB, which better balances the parameter distribution and enhances the capture of high-level features. The incorporation of residual learning enables the network to focus on reconstructing crucial image details, thereby improving both the model’s capacity and the accuracy of detail restoration.

\subsection{Loss Function}
\label{sec:II-D}
To improve frame detail and structural accuracy, we use a hybrid loss function combining L1 loss and Multi-Scale Structural Similarity Index Measure (MS-SSIM). And to further retention of high-frequency details, we employ frequency constraints to regularize network training,the loss function as:
\begin{equation}
    L_{\text{spa}} = \alpha \| \hat{I}_t - I_t \|_1 + (1 - \alpha)(1 - \text{MS-SSIM}(\hat{I}_t, I_t)),
\end{equation}
\begin{equation}
    L_{\text{fft}} = \| \mathcal{F}(\hat{I}_t) - \mathcal{F}(I_t) \|_1,
\end{equation}
\begin{equation}
    L_{\text{total}} = L_{\text{spa}} + \mu L_{\text{fft}}.
\end{equation}
Here, \( \hat{I}_t \) and \( I_t \) represent the reconstructed and original frames, respectively. The symbol \(\|\cdot\|_1\) denotes the \(L_1\)-norm, and \( \mathcal{F} \) represents the Fast Fourier Transform (FFT). Additionally, \( \alpha \) and \( \mu \) are weight parameters, empirically set to 0.7 and 70, respectively.

\section{Experimental results and discussion}
\subsection{DataSets and Implementations}
We evaluated our approach using the Bunny \cite{bunny}, UVG \cite{uvg}, and DAVIS \cite{davis} datasets. The UVG dataset consists of seven video sequences, each with 300 or 600 frames at a resolution of 1080$\times$1920. 
For the DAVIS dataset, we used the validation set, which includes 20 videos, each with a resolution of 1080$\times$1920.
The decoder stride was set to [5, 2, 2, 2, 2] for the Bunny dataset and [5, 2, 2, 2, 2, 2] for UVG and DAVIS. For evaluation, we used Peak Signal-to-Noise Ratio (PSNR) and Multi-Scale Structural Similarity Index Measure (MS-SSIM) to assess distortion and measured the video compression bit rate in bits per pixel (bpp). Our method was benchmarked against four NeRV-based approaches with publicly available codes: NeRV \cite{nerv}, E-NeRV \cite{E-nerv}, HNeRV \cite{hnerv}, and boosted HNeRV (Boost) \cite{boost}. To ensure a fair comparison, we adjusted the number of channels in each method to maintain consistent overall capacity across the experimental models. Training was performed using the Adan optimizer \cite{adan} with a cosine learning rate decay, starting from an initial learning rate of \(3 \times 10^{-3}\). The batch size was set to 1. All experiments were implemented in PyTorch and executed on a single NVIDIA GTX 4090 GPU, with a model size of 3M and 300 training epochs unless otherwise specified.

\subsection{Results}
\subsubsection{Video regression}      
\begin{figure}[b]
\centering
    \includegraphics[width=1\linewidth]{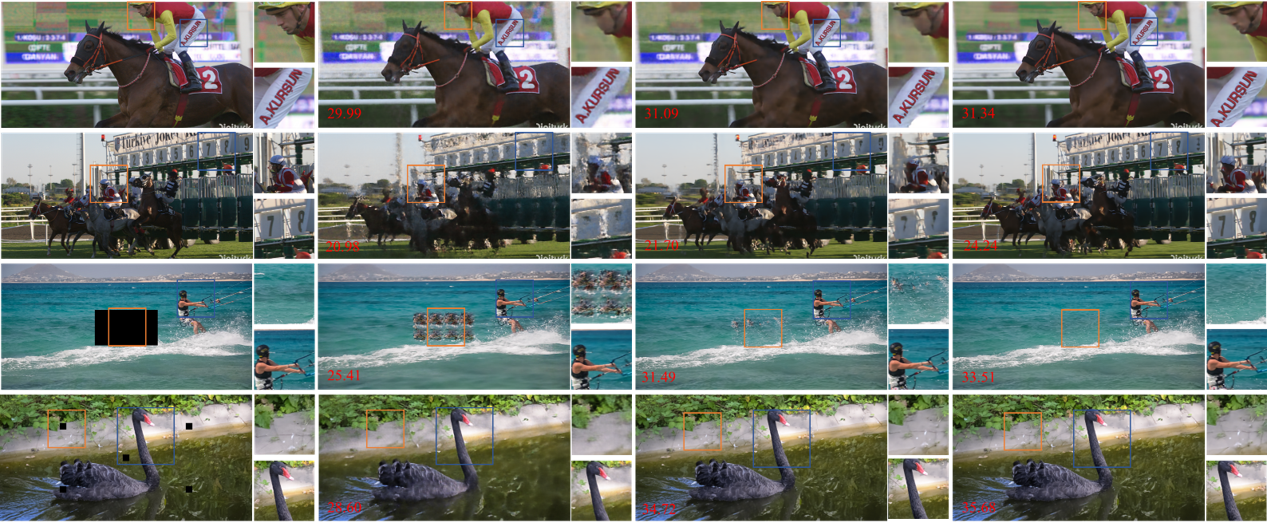}
\caption{The visualization comparison results are arranged in a top-to-bottom format, highlighting video reconstruction, interpolation, central inpainting (Mask-C), and dispersed inpainting (Mask-S) tasks on the DAVIS validation and UVG datasets.  The first column shows the ground truth, followed by the baseline results from NeRV, HNeRV, and our method.  The red numbers represent the corresponding PSNR values.
}
\label{fig:visualization}
\end{figure}

We evaluated the video regression performance of our method against other approaches on the Bunny and UVG datasets. As shown in Table~\ref{tab:bunny_performance}, our method consistently achieved superior video reconstruction quality on the Bunny dataset across various model sizes. Additionally, performance analysis over training iterations reveals that our method converges faster and sustains higher performance throughout training. For the UVG dataset, the results presented in Table~\ref{tab:uvg_performance} demonstrate that our method attained the highest average performance across all video. Notably, it improved the average PSNR by 0.64 dB at a resolution of \(960 \times 1920\) and by 0.22 dB at \(480 \times 960\) compared to the next best method. Fig.~\ref{fig:visualization} (first row) illustrates our method’s ability to preserve fine video details, such as reconstructing text and facial features more effectively than competing approaches. These results underscore the significant impact of the frequency enhancement strategy in improving frame reconstruction quality, highlighting the superiority and robustness of our approach.

\begin{table}[t]
\caption{PSNR with varying model size and epochs on Bunny.}
\begin{center}
\setlength{\tabcolsep}{4pt}        
\renewcommand{\arraystretch}{1.1}  
\begin{tabular}{|c|ccc|ccc|}
\hline
\multirow{2}{*}{\textbf{Method}} 
& \multicolumn{3}{c|}{\textbf{Model Size}} 
& \multicolumn{3}{c|}{\textbf{Epoch}} \\
\cline{2-7}
& \textbf{0.75M} & \textbf{1.5M} & \textbf{3M}
& \textbf{300} & \textbf{600} & \textbf{1200} \\
\hline
NeRV       & 28.46 & 30.87 & 33.21 & 33.21 & 34.47 & 35.07 \\
E-NeRV     & 30.95 & 32.09 & 36.72 & 36.72 & 38.20 & 39.48 \\
HNeRV      & 32.18 & 35.19 & 37.43 & 37.43 & 39.36 & 40.02 \\
Boost & 35.53 & 38.95 & 41.50 & 41.50 & 42.03 & 42.34 \\
\hline
Ours     & \textbf{35.82} & \textbf{39.10} & \textbf{41.82} & \textbf{41.82} & \textbf{42.40} & \textbf{42.73} \\
\hline
\end{tabular}
\end{center}
\label{tab:bunny_performance}
\end{table} 

\begin{table*}[htbp]
\caption{Video Regression Performance in terms of PSNR and MS-SSIM on Different Resolution UVG Dataset}
\begin{center}
\begin{tabular}{|c|c|c|c|c|c|c|c|c|c|}
\hline
\textbf{Resolution} & \textbf{Method} & \textbf{Beauty} & \textbf{Honey.} & \textbf{Bosph.} & \textbf{Yacht.} & \textbf{Ready.} & \textbf{Jockey} & \textbf{Shake.} & \textbf{Avg.} \\
\hline
\multirow{5}{*}{1920×960} & NeRV   & 33.25/0.8886 & 37.26/0.9794 & 33.22/0.9305 & 28.03/0.8726 & 24.84/0.8310 & 31.74/0.8874 & 33.08/0.9325 & 31.63/0.9031 \\
                          & E-NeRV & 33.53/0.8958 & 39.04/0.9845 & 33.81/0.9442 & 27.74/0.8951 & 24.09/0.8515 & 29.35/0.8805 & 34.54/0.9467 & 31.73/0.9140 \\
                          & Hnerv  & 33.58/0.8941 & 38.96/0.9844 & 34.73/0.9451 & 29.26/0.8907 & 25.74/0.8420 & 32.04/0.8802 & 34.57/0.9450 & 32.69/0.9116 \\
                          & Boost  & 33.93/0.9006 & 39.62/0.9854 & 36.00/0.9652 & 29.69/0.9079 & 28.33/0.9173 & 34.51/0.9326 & 35.89/0.9581 & 34.00/0.9382 \\
\cline{2-10}
                          & Ours & \textbf{34.11/0.9032} & \textbf{39.69/0.9854} & \textbf{37.09/0.9729} & \textbf{30.58/0.9230} & \textbf{29.68/0.9367} & \textbf{35.36/0.9438} & \textbf{35.97/0.9589} & \textbf{34.64/0.9463} \\
\hline
\multirow{5}{*}{960×480}  & NeRV   & 32.38/0.9346 & 36.64/0.9912 & 32.95/0.9577 & 28.07/0.9183 & 24.55/0.8884 & 31.33/0.9154 & 32.74/0.9603 & 31.24/0.9380 \\
                          & E-NeRV & 32.59/0.9399 & 38.47/0.9936 & 33.72/0.9705 & 27.86/0.9393 & 24.05/0.9069 & 28.98/0.9084 & 34.06/0.9715 & 31.39/0.9472 \\
                          & Hnerv  & 32.81/0.9341 & 38.52/0.9936 & 34.58/0.9703 & 29.24/0.9354 & 25.73/0.9112 & 32.04/0.9151 & 34.34/0.9698 & 32.47/0.9470 \\
                          & Boost  & 33.06/0.9437 & 38.96/0.9942 & 36.41/0.9843 & 30.10/0.9512 & 28.80/0.9603 & 34.29/0.9570 & 35.25/0.9768 & 33.84/0.9668 \\
\cline{2-10}
                          & Ours & \textbf{33.48/0.9462} & \textbf{39.17/0.9958} & \textbf{36.64/0.9858} & \textbf{30.40/0.9557} & \textbf{29.28/0.9693} & \textbf{34.83/0.9647} & \textbf{35.81/0.9793} & \textbf{34.23/0.9710} \\
\hline
\end{tabular}
\label{tab:uvg_performance}
\end{center}
\end{table*}

\subsubsection{Video Compression}
\begin{figure}[htbp]
    \centering
    \begin{minipage}[t]{0.24\textwidth} 
        \centering
        \includegraphics[width=\textwidth]{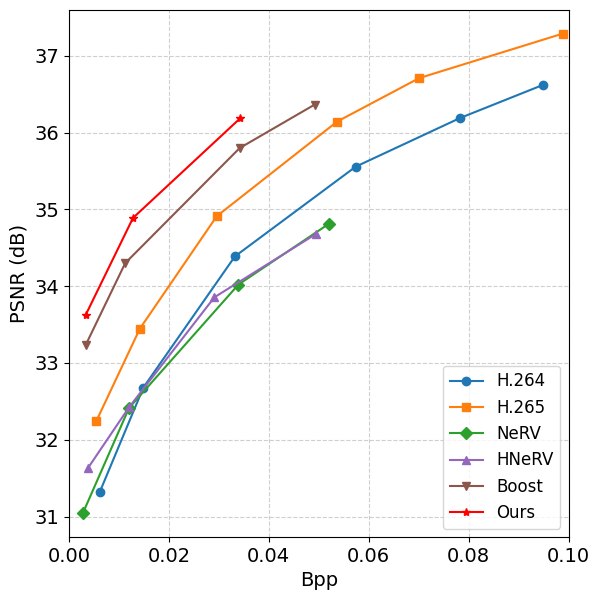}
        \label{fig:psnr}
    \end{minipage}
    \begin{minipage}[t]{0.24\textwidth} 
        \centering
        \includegraphics[width=\textwidth]{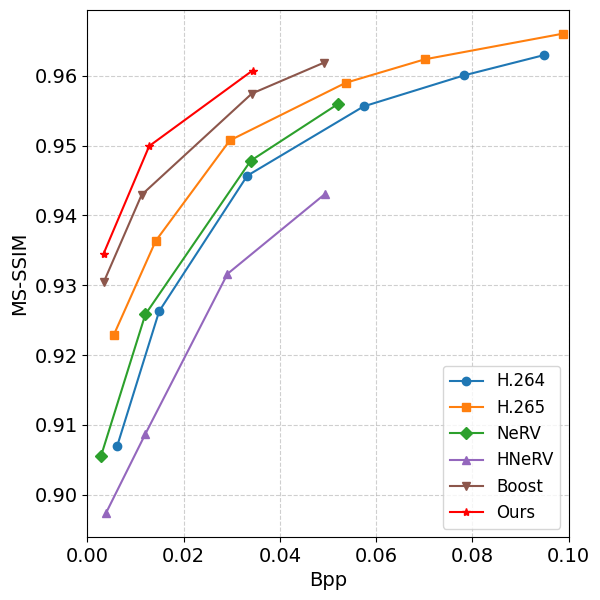}
        \label{fig:msssim}
    \end{minipage}
    \caption{The rate-distortion curve on UVG in terms of PSNR and MS-SSIM.} 
    \label{fig:comparison}
\end{figure}

We employed the consistent entropy minimization method \cite{boost} for model compression training. Each model was fine-tuned for 100 iterations using an initial learning rate of \(5 \times 10^{-4}\) with a cosine decay schedule. Our approach was benchmarked against two traditional codecs, H.264 \cite{h264} and H.265 \cite{h265}, using FFmpeg with the "very slow" presets. Rate-distortion (R-D) curves, evaluated using PSNR and MS-SSIM metrics on the average of the UVG dataset, are shown in Fig.~\ref{fig:comparison}. The results indicate that, at equivalent bit rates, our method consistently surpasses both traditional codecs and other NeRV-based approaches, achieving superior compression efficiency.

\subsubsection{video Interpolation}
We evaluated the video interpolation performance of our method on the UVG dataset. Even frames were used for training, while odd frames served as the test set. Table~\ref{tab:uvg_interpolation} presents the quantitative results, showing that our method achieved the highest PSNR scores across most video sequences. The qualitative results, illustrated in Fig.~\ref{fig:visualization} (second row), demonstrates that only our method successfully reconstructed the gate numbers on the horse racing starting gates with high fidelity. These results indicate that our method demonstrated superior performance compared to competing approaches in terms of overall interpolation quality.
\begin{table}[hbp]
\caption{Video interpolation results on UVG dataset in PSNR.}
\begin{center}
\setlength{\tabcolsep}{3pt} 
\renewcommand{\arraystretch}{0.9} 
\begin{tabular}{|c| c c c c c c c|}
\hline
\textbf{Method} & \textbf{Beauty} & \textbf{Bosph.} & \textbf{Honey.} & \textbf{Jockey} & \textbf{Ready.} & \textbf{Yacht.} & \textbf{Shake.} \\
\hline
NeRV    & 31.26 & 32.21 & 36.84 & 22.24 & 20.05 & 26.09 & 32.09 \\
E-NeRV  & 31.25 & 33.36 & 38.62 & 22.35 & 20.08 & 26.74 & 32.82 \\
Hnerv   & 31.42 & 34.00 & 39.07 & 23.02 & 20.71 & 26.74 & 32.58 \\
Boost   & 31.59 & 35.92 & \textbf{39.32} & 22.95 & 21.34 & \textbf{27.98} & 32.65 \\
\hline
Ours  & \textbf{31.67} & \textbf{36.48} & 39.22 & \textbf{23.56} & \textbf{22.43} & 27.87 & \textbf{32.92} \\
\hline
\end{tabular}
\label{tab:uvg_interpolation}
\end{center}
\end{table}

\begin{table}[htbp]
\caption{VIDEO INPAINTING RESULTS ON DAVIS DATASET IN PSNR.}
\begin{center}
\setlength{\tabcolsep}{1.8pt}  
\renewcommand{\arraystretch}{0.9}  
\begin{tabular}{|c|ccccccccc|}
\hline
\textbf{Method} 
& Black. & Break. & Camel & Car-r. & Car-s. & Goat & Kite. & Libby & Para. \\
\hline
\multicolumn{10}{|c|}{\textbf{Mask-C}} \\
\hline
NeRV & 24.11 & 20.16 & 21.21 & 21.24 & 23.07 & 22.03 & 23.92 & 25.71 & 25.95 \\
E-NeRV & 26.38 & 22.15 & \textbf{22.62} & 22.73 & 23.21 & 23.43 & 26.71 & 26.91 & 26.65 \\
HNeRV & 26.45 & 20.23 & 17.74 & 21.71 & 21.05 & 23.06 & 24.73 & 23.39 & 26.00 \\
Boost & 29.18 & 20.24 & 19.81 & 22.36 & 23.65 & 25.10 & 27.49 & 26.96 & 28.07 \\ \hline
Ours & \textbf{29.52} & \textbf{20.61} & 21.30 & \textbf{22.41} & \textbf{23.69} & \textbf{25.22} & \textbf{27.71} & \textbf{27.73} & \textbf{28.68} \\
\hline
\multicolumn{10}{|c|}{\textbf{Mask-S}} \\
\hline
NeRV & 27.06 & 25.48 & 23.70 & 23.92 & 26.58 & 24.04 & 29.34 & 29.81 & 29.03 \\
E-NeRV & 29.53 & 26.97 & 25.70 & 26.32 & 30.63 & 25.34 & 32.87 & 31.39 & 30.62 \\
HNeRV & 30.20 & 26.34 & 26.13 & 28.64 & 31.01 & 26.91 & 33.49 & 28.66 & 30.99 \\ 
Boost & 34.10 & \textbf{33.10} & 31.08 & 31.90 & 35.85 & 30.59 & 37.08 & 37.35 & 33.64 \\ \hline
Ours & \textbf{35.30} & 32.78 & \textbf{31.87} & \textbf{32.26} & \textbf{36.58} & \textbf{31.33} & \textbf{37.26} & \textbf{38.27} & \textbf{34.54} \\
\hline
\end{tabular}
\label{tab:inpainting_performance}
\end{center}
\end{table}
\begin{table*}[htbp]
\centering
\caption{Ablation Study of FANeRV On Bunny Dataset, with results presented in PSNR and MS-SSIM.}
\resizebox{\textwidth}{!}{%
\begin{tabular}{|c|c|c|c|c|c|c|c|c|c|}
\hline
\textbf{Metric}         & \textbf{Ours} & \textbf{w/o WFUB} & \textbf{w/o FSFB} & \textbf{w/o TGFN} & \textbf{w/o CREB} & \textbf{w/ Large-kernels conv} & \textbf{w/ Self-attention} & \textbf{w/ ConvTranspose} & \textbf{w/ Bilinear} \\ \hline
\textbf{PSNR}           & \textbf{41.82}         & 41.32             & 41.44             & 41.58             & 41.51           & 41.72                         & 41.65                     & 41.57                     & 40.89               \\ \hline
\textbf{SSIM}           & \textbf{0.9942}        & 0.9933            & 0.9936            & 0.9937            & 0.9939          & 0.9940                        & 0.9935                    & 0.9938                    & 0.9929              \\ \hline
\end{tabular}%
}
\label{tab:ablation}
\end{table*}

\subsubsection{Video Inpainting}
We evaluated the performance of FANeRV on video inpainting tasks using the DAVIS validation dataset. We conducted both disperse (Mask-S) and central masking (Mask-C) experiments. In the disperse masking setup, five 50$\times$50 regions were masked in each video frame during training. For the central masking experiment, a region covering one-quarter of the video's width and height was masked. The training goal was to reconstruct the complete video frame. Table \ref{tab:inpainting_performance} and Fig.\ref{fig:visualization} present the quantitative and qualitative results, showing that our method's robust structural design significantly improved the recovery of masked regions compared to previous methods.

\begin{figure}[htbp]
    \centering
\includegraphics[width=0.7\linewidth]{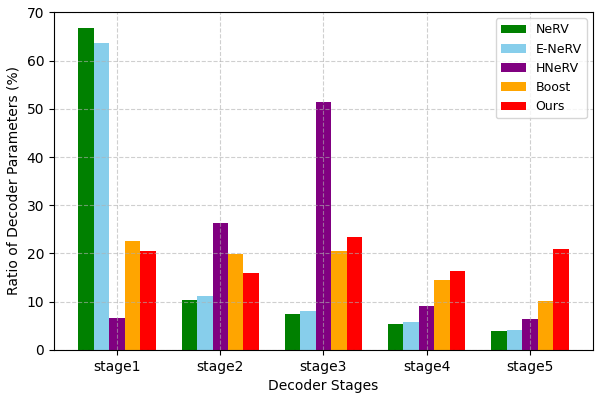}

    \caption{Parameter distribution across decoder blocks in various models.}
    \label{fig:Parameter}
\end{figure}
\subsubsection{Ablation Study}
Tab.\ref{tab:ablation} demonstrates the effectiveness of each component in the proposed FANeRV. Multiple variants of the original model were generated and trained on the Bunny dataset to ensure fair comparisons. Initially, individual modules (WFUB, FSFB, TGFN, and CREB) were disabled to evaluate their contributions by comparing their performance to the baseline FANeRV. The results in Table~\ref{tab:ablation} show significant drops in PSNR values, highlighting the critical role of each module. Additionally, we visualized the decoder parameter distribution at different stages in Fig.~\ref{fig:Parameter}, showing that our block achieves a more uniform parameter distribution and that the increased parameters in later stages aid in finer detail recovery. Furthermore, we replaced the multi-resolution deep feature modulation branch with large convolutional kernels and channel-wise self-attention mechanisms. This modification led to reduced PSNR values, demonstrating that our approach effectively captures long-range dependencies while maintaining computational efficiency. To further evaluate the multi-resolution fusion strategy, we replaced it with transposed convolution and bilinear upsampling methods. The results confirm that our fusion strategy delivers superior performance. These ablation studies underline the importance of each module in FANeRV and validate the robustness of its overall design.

\section{Conclusion}
In this paper, we introduce Frequency Separation and Augmentation based Neural Representation for video, which explicitly separates high and low-frequency information and develops distinct recovery methods tailored to their specific characteristics. Extensive experimental results show that FANeRV outperforms several typical methods in reconstruction performance while maintaining the same model capacity.



\bibliographystyle{IEEEbib}
\bibliography{icme2025}

\begin{thebibliography}{10}

\bibitem{nerf}
Ben Mildenhall, Pratul~P Srinivasan, Matthew Tancik, Jonathan~T Barron, Ravi Ramamoorthi, and Ren Ng,
\newblock ``Nerf: Representing scenes as neural radiance fields for view synthesis,''
\newblock {\em Communications of the ACM}, vol. 65, no. 1, pp. 99--106, 2021.

\bibitem{siren}
Vincent Sitzmann, Julien Martel, Alexander Bergman, David Lindell, and Gordon Wetzstein,
\newblock ``Implicit neural representations with periodic activation functions,''
\newblock {\em Advances in neural information processing systems}, vol. 33, pp. 7462--7473, 2020.

\bibitem{3d1}
Vincent Sitzmann, Michael Zollh{\"o}fer, and Gordon Wetzstein,
\newblock ``Scene representation networks: Continuous 3d-structure-aware neural scene representations,''
\newblock {\em Advances in Neural Information Processing Systems}, vol. 32, 2019.

\bibitem{nerv}
Hao Chen, Bo~He, Hanyu Wang, Yixuan Ren, Ser~Nam Lim, and Abhinav Shrivastava,
\newblock ``Nerv: Neural representations for videos,''
\newblock {\em Advances in Neural Information Processing Systems}, vol. 34, pp. 21557--21568, 2021.

\bibitem{coin}
Emilien Dupont, Adam Goli{\'n}ski, Milad Alizadeh, Yee~Whye Teh, and Arnaud Doucet,
\newblock ``Coin: Compression with implicit neural representations,''
\newblock {\em arXiv preprint arXiv:2103.03123}, 2021.

\bibitem{E-nerv}
Zizhang Li, Mengmeng Wang, Huaijin Pi, Kechun Xu, Jianbiao Mei, and Yong Liu,
\newblock ``E-nerv: Expedite neural video representation with disentangled spatial-temporal context,''
\newblock in {\em European Conference on Computer Vision}. Springer, 2022, pp. 267--284.

\bibitem{hnerv}
Hao Chen, Matthew Gwilliam, Ser-Nam Lim, and Abhinav Shrivastava,
\newblock ``Hnerv: A hybrid neural representation for videos,''
\newblock in {\em Proceedings of the IEEE/CVF Conference on Computer Vision and Pattern Recognition}, 2023, pp. 10270--10279.

\bibitem{boost}
Xinjie Zhang, Ren Yang, Dailan He, Xingtong Ge, Tongda Xu, Yan Wang, Hongwei Qin, and Jun Zhang,
\newblock ``Boosting neural representations for videos with a conditional decoder,''
\newblock in {\em Proceedings of the IEEE/CVF Conference on Computer Vision and Pattern Recognition (CVPR)}, June 2024, pp. 2556--2566.

\bibitem{spectralbias}
Nasim Rahaman, Aristide Baratin, Devansh Arpit, Felix Draxler, Min Lin, Fred Hamprecht, Yoshua Bengio, and Aaron Courville,
\newblock ``On the spectral bias of neural networks,''
\newblock in {\em International Conference on Machine Learning}. PMLR, 2019, pp. 5301--5310.

\bibitem{vqnerv}
Yunjie Xu, Xiang Feng, Feiwei Qin, Ruiquan Ge, Yong Peng, and Changmiao Wang,
\newblock ``Vq-nerv: A vector quantized neural representation for videos,''
\newblock {\em arXiv preprint arXiv:2403.12401}, 2024,
\newblock unpublished.

\bibitem{qsnerv}
Chang Wu, Guancheng Quan, Gang He, Xin-Quan Lai, Yunsong Li, Wenxin Yu, Xianmeng Lin, and Cheng Yang,
\newblock ``Qs-nerv: Real-time quality-scalable decoding with neural representation for videos,''
\newblock in {\em Proceedings of the 32nd ACM International Conference on Multimedia}, 2024, pp. 2584--2592.

\bibitem{snerv}
Jina Kim, Jihoo Lee, and Je-Won Kang,
\newblock ``Snerv: Spectra-preserving neural representation for video,''
\newblock in {\em European Conference on Computer Vision}. Springer, 2025, pp. 332--348.

\bibitem{nerv++}
Ahmed Ghorbel, Wassim Hamidouche, and Luce Morin,
\newblock ``Nerv++: An enhanced implicit neural video representation,''
\newblock {\em arXiv preprint arXiv:2402.18305}, 2024,
\newblock unpublished.

\bibitem{convnet}
Zhuang Liu, Hanzi Mao, Chao-Yuan Wu, Christoph Feichtenhofer, Trevor Darrell, and Saining Xie,
\newblock ``A convnet for the 2020s,''
\newblock in {\em Proceedings of the IEEE/CVF conference on computer vision and pattern recognition}, 2022, pp. 11976--11986.

\bibitem{crnet}
Kangzhen Yang, Tao Hu, Kexin Dai, Genggeng Chen, Yu~Cao, Wei Dong, Peng Wu, Yanning Zhang, and Qingsen Yan,
\newblock ``Crnet: A detail-preserving network for unified image restoration and enhancement task,''
\newblock {\em arXiv preprint arXiv:2404.14132}, 2024,
\newblock unpublished.

\bibitem{asymmetric}
Kai Hu, Yu~Liu, Fang Xu, Renhe Liu, Han Wang, and Shenghui Song,
\newblock ``Asymmetric neural image compression with high-preserving information,''
\newblock in {\em 2024 IEEE International Symposium on Circuits and Systems (ISCAS)}. IEEE, 2024, pp. 1--5.

\bibitem{afftransformer}
Zhisheng Lu, Juncheng Li, Hong Liu, Chaoyan Huang, Linlin Zhang, and Tieyong Zeng,
\newblock ``Transformer for single image super-resolution,''
\newblock in {\em Proceedings of the IEEE/CVF conference on computer vision and pattern recognition}, 2022, pp. 457--466.

\bibitem{harr}
Ingrid Daubechies,
\newblock ``The wavelet transform, time-frequency localization and signal analysis,''
\newblock {\em IEEE transactions on information theory}, vol. 36, no. 5, pp. 961--1005, 1990.

\bibitem{hendrycks2016gaussian}
Dan Hendrycks and Kevin Gimpel,
\newblock ``Gaussian error linear units (gelus),''
\newblock {\em arXiv preprint arXiv:1606.08415}, 2016,
\newblock unpublished.

\bibitem{bunny}
Ton Roosendaal,
\newblock ``Big buck bunny,''
\newblock in {\em ACM SIGGRAPH ASIA 2008 computer animation festival}, pp. 62--62. 2008.

\bibitem{uvg}
Alexandre Mercat, Marko Viitanen, and Jarno Vanne,
\newblock ``Uvg dataset: 50/120fps 4k sequences for video codec analysis and development,''
\newblock in {\em Proceedings of the 11th ACM Multimedia Systems Conference}, 2020, pp. 297--302.

\bibitem{davis}
Federico Perazzi, Jordi Pont-Tuset, Brian McWilliams, Luc Van~Gool, Markus Gross, and Alexander Sorkine-Hornung,
\newblock ``A benchmark dataset and evaluation methodology for video object segmentation,''
\newblock in {\em Proceedings of the IEEE conference on computer vision and pattern recognition}, 2016, pp. 724--732.

\bibitem{adan}
Xingyu Xie, Pan Zhou, Huan Li, Zhouchen Lin, and Shuicheng Yan,
\newblock ``Adan: Adaptive nesterov momentum algorithm for faster optimizing deep models,''
\newblock {\em arXiv preprint arXiv:2208.06677}, 2022.

\bibitem{h264}
Thomas Wiegand, Gary~J Sullivan, Gisle Bjontegaard, and Ajay Luthra,
\newblock ``Overview of the h. 264/avc video coding standard,''
\newblock {\em IEEE Transactions on circuits and systems for video technology}, vol. 13, no. 7, pp. 560--576, 2003.

\bibitem{h265}
Gary~J Sullivan, Jens-Rainer Ohm, Woo-Jin Han, and Thomas Wiegand,
\newblock ``Overview of the high efficiency video coding (hevc) standard,''
\newblock {\em IEEE Transactions on circuits and systems for video technology}, vol. 22, no. 12, pp. 1649--1668, 2012.

\end{thebibliography}

\end{document}